\documentclass[conference]{IEEEtran}
\IEEEoverridecommandlockouts
\usepackage{cite}
\usepackage{amsmath,amssymb,amsfonts}
\usepackage{algorithmic}
\usepackage{graphicx}
\usepackage{textcomp}
\usepackage{xcolor}
\usepackage{booktabs}
\usepackage{multirow}
\usepackage{lipsum}
\usepackage{enumitem}
\usepackage{url}
\usepackage{hyperref}
\usepackage[hyphenbreaks]{breakurl}
\usepackage{algorithm}

\usepackage[T1]{fontenc}
\usepackage[utf8]{inputenc}

\newcommand{\hindiuniversity}{vishwavidyalaya (university)}
\newcommand{\hindiworld}{vishwa (world)}
\newcommand{\hindischool}{vidyalaya (school)}
\newcommand{\hindiwaterconservation}{jal sanrakshan (water conservation)}
\newcommand{\hindisavingwater}{pani bachana (saving water)}
\newcommand{\hindiwatermanagement}{jal sansadhan prabandhan (water management)}

\def\BibTeX{{\rm B\kern-.05em{\sc i\kern-.025em b}\kern-.08em
    T\kern-.1667em\lower.7ex\hbox{E}\kern-.125emX}}
\begin{document}

\title{DeepRAG: Building a Custom Hindi Embedding Model for Retrieval Augmented Generation from Scratch}

\author{\IEEEauthorblockN{Nandakishor M}
    \IEEEauthorblockA{
    Deepmost Innovations
    }
}

\maketitle

\begin{abstract}
While Large Language Models (LLMs) excel in text generation, their performance in Retrieval Augmented Generation (RAG) systems heavily depends on the quality of text embeddings. For non-English languages like Hindi, the lack of high-quality, dedicated embedding models remains a significant challenge. This paper presents DeepRAG, a comprehensive framework for developing custom Hindi-specific text embeddings from scratch for RAG applications. We detail our end-to-end process, including corpus collection from diverse Hindi sources, specialized tokenizer training with SentencePiece, transformer architecture design with semantic pooling strategies, and model training with contrastive learning techniques. Our evaluation demonstrates that DeepRAG embeddings significantly outperform multilingual alternatives in Hindi semantic similarity tasks, with a 23\% improvement in retrieval precision. We further demonstrate the integration of our embeddings with LangChain for building effective Hindi RAG systems. The detailed methodology provides a roadmap for creating domain-specific embeddings for low-resource languages, addressing critical gaps in multilingual NLP infrastructure.
\end{abstract}


\section{Introduction}
\label{sec:introduction}

The advent of large language models (LLMs) has revolutionized natural language processing capabilities, enabling sophisticated text generation across domains. However, these models face challenges with factuality, knowledge limitations, and hallucinations \cite{gao2023hallucination}. Retrieval Augmented Generation (RAG) offers a solution by retrieving relevant documents before generation, enhancing both accuracy and factuality.

A critical, often overlooked component in the RAG pipeline is the quality of text embeddings used for retrieval. While numerous embedding models exist for English, high-quality embeddings for other languages remain limited. This gap is particularly pronounced for Hindi, one of the world's most widely spoken languages.

Existing multilingual embedding solutions like multilingual-E5 \cite{wang2022text} and LaBSE \cite{feng2022language} provide some coverage for Hindi, but these models sacrifice language-specific performance for multilingual capabilities. They typically under-represent Hindi semantic nuances, resulting in suboptimal retrieval performance when applied to Hindi RAG systems.

In this paper, I present DeepRAG, a complete framework for building custom, high-performance Hindi embeddings specifically designed for RAG applications. Rather than fine-tuning existing models, DeepRAG was built entirely from scratch—from corpus collection and tokenizer training to model architecture design and optimization. This ground-up approach allowed us to incorporate language-specific considerations throughout the development process.

The key contributions of this work include:

\begin{itemize}
    \item A comprehensive methodology for building language-specific embedding models from scratch, with each component optimized for Hindi.
    \item A specialized SentencePiece tokenizer trained on a diverse corpus of over 2.7 million Hindi texts, incorporating linguistic-aware subword segmentation.
    \item A custom transformer architecture with enhanced attention mechanisms and pooling strategies specifically designed for Hindi semantic representation.
    \item A multi-stage training process using contrastive learning and synthetic data generation for robust embeddings.
    \item Integration techniques with LangChain for building end-to-end Hindi RAG systems.
    \item Extensive comparative evaluation demonstrating significant improvements over multilingual approaches.
\end{itemize}

Our findings show that a dedicated, language-specific approach yielded substantial gains in embedding quality, with DeepRAG embeddings demonstrating a 23\% improvement in retrieval precision over the best multilingual alternatives. I believe our methodology provides valuable insights for developing similar solutions for other low-resource languages where general-purpose multilingual models fall short.

\section{Related Work}
\label{sec:related_work}

\subsection{Multilingual Embedding Models}
Several established multilingual embedding models currently support Hindi to varying degrees. Multilingual-E5 \cite{wang2022text} covers 100+ languages including Hindi, while LaBSE (Language-agnostic BERT Sentence Embeddings) \cite{feng2022language} was designed specifically for cross-lingual alignment across 109 languages.

These models, while impressive in their breadth, face inherent trade-offs. The "curse of multilinguality" \cite{conneau2020unsupervised} highlights how performance on individual languages degrades as more languages are added to a model. In my personal testing with Hindi documents, I've found this degradation to be particularly pronounced for semantic search applications.

\subsection{Language-Specific Embedding Models}
Language-specific embedding models have demonstrated significant performance improvements over multilingual alternatives. English-specific models like MPNet \cite{song2020mpnet} and E5 \cite{wang2022text} consistently outperform their multilingual counterparts in English tasks. Similarly, BGE \cite{xiao2023c} for Chinese and KUZU \cite{park2022klue} for Korean have shown the value of language-dedicated approaches.

For Hindi, however, the landscape remains surprisingly sparse. Work by Kakwani et al. \cite{kakwani2020indicnlpsuite} introduced IndicBERT, but focused primarily on classification rather than embedding quality. Kumar et al. \cite{kumar2022hindi} explored Hindi word embeddings, but comprehensive sentence embedding models remain notably absent.

\subsection{RAG Systems and Embedding Quality}
Recent work by Lewis et al. \cite{lewis2020retrieval} introduced the RAG paradigm, while Gao et al. \cite{gao2023retrieval} emphasized the critical impact of embedding quality on RAG performance. Their findings indicate that retrieval quality is often the primary limiting factor in RAG systems.

Through my experiments with Hindi RAG, I've consistently found that existing embedding models fail to capture the nuanced semantic relationships in Hindi documents. This observation aligns with findings from Ruder et al. \cite{ruder2021unsupervised}, who noted that language-specific pretraining significantly improves downstream task performance.

\section{Corpus Collection and Analysis}
\label{sec:corpus}

The foundation of any high-quality embedding model is a diverse and representative text corpus. For Hindi, this presented unique challenges due to the relative scarcity of digitized content compared to languages like English.

\subsection{Data Sources and Collection}
I prioritized diversity of both content and style to ensure robust embeddings. Our final corpus comprised text from the following sources:

\begin{itemize}
    \item IITB Parallel Corpus: 1.2 million Hindi sentences from various domains
    \item Samanantar: 750,000 samples of general Hindi text
    \item Oscar Hindi: 450,000 sentences from web crawls
    \item CC-100 Hindi: 300,000 sentences of web content
    \item Hindi Wikipedia: 150,000 articles covering encyclopedic knowledge
    \item Hindi news articles: 100,000 news pieces covering current events
    \item XNLI Hindi: 50,000 premise-hypothesis pairs useful for semantic reasoning
    \item IndicGLUE: 30,000 samples from diverse tasks
    \item Hindi literature: 5,000 passages from classic and modern Hindi literature
\end{itemize}

Several of these datasets required custom extraction scripts to isolate the Hindi portions and remove non-Hindi text contamination. After collection, the raw corpus totaled approximately 3.1 million text samples.

\subsection{Corpus Cleaning and Preprocessing}
Raw datasets contained significant noise, including code-mixing with English, HTML artifacts, and irregular Unicode representations. Our cleaning pipeline included:

\begin{enumerate}
    \item Unicode normalization (NFC form)
    \item Removal of non-Hindi text segments via Unicode range filtering
    \item Deduplication through exact and near-duplicate detection
    \item Length filtering (removing very short or excessively long texts)
    \item Specialized Hindi normalization using IndicNLP's normalizer
    \item Removal of frequent patterns indicating low-quality content
\end{enumerate}

The cleaning process reduced our corpus to approximately 2.7 million high-quality Hindi text samples. An example of our cleaning function is shown below:

\begin{algorithm}
\caption{Hindi Text Cleaning}
\begin{algorithmic}[1]
\STATE \textbf{Input:} Raw Hindi text $T$
\STATE \textbf{Output:} Cleaned text $T'$
\STATE $T \gets \text{strip}(T)$
\STATE $T \gets \text{replace\_urls\_with\_placeholder}(T)$
\STATE $T \gets \text{unicode\_normalize\_NFC}(T)$
\STATE $T \gets \text{hindi\_normalizer.normalize}(T)$
\STATE $T \gets \text{remove\_repeating\_chars}(T)$
\STATE $T \gets \text{replace\_numbers\_with\_placeholder}(T)$
\STATE $T \gets \text{normalize\_whitespace}(T)$
\RETURN $T'$
\end{algorithmic}
\end{algorithm}

\subsection{Corpus Analysis}
To inform our tokenizer and model design decisions, I conducted extensive analysis on the cleaned corpus, examining:

\begin{itemize}
    \item Character and word frequency distributions
    \item Sentence length statistics
    \item Common word collocations
    \item Subword unit effectiveness through various segmentation approaches
    \item Topic diversity via clustering analysis
\end{itemize}

This analysis revealed several insights specific to Hindi that informed our approach:

\begin{enumerate}
    \item Hindi's agglutinative features create long compound words that benefit from subword tokenization
    \item The frequent absence of spaces between certain Hindi words requires careful tokenization
    \item Common transliterations from English technical terms require special handling
    \item Sanskrit-derived vocabulary follows patterns that benefit from linguistically-informed tokenization
\end{enumerate}

Compared to English, Hindi demonstrates a higher ratio of unique characters to total characters (0.0089 vs 0.0024 for English) and a lower ratio of unique words to total words (0.053 vs 0.071 for English), suggesting different optimal tokenization strategies.

\section{Custom SentencePiece Tokenizer for Hindi}
\label{sec:tokenizer}

While most embedding projects start with existing tokenizers, I found that Hindi's linguistic features warranted a specialized approach.

\subsection{Tokenizer Design Considerations}
General-purpose tokenizers like those from BERT or RoBERTa typically perform poorly on Hindi due to:

\begin{itemize}
    \item Under-representation of Hindi Unicode characters
    \item Incorrect segmentation of compound words
    \item Poor handling of Hindi-specific punctuation
    \item Inefficient tokenization of common Hindi morphological patterns
\end{itemize}

I experimented with several tokenization approaches including BPE, WordPiece, and Unigram, ultimately finding that the Unigram model from SentencePiece performed best for Hindi when properly configured.

\subsection{Tokenizer Training Process}

The tokenizer training process required several key design decisions:

\begin{enumerate}
    \item \textbf{Vocabulary Size}: Through ablation studies, I determined that a vocabulary size of 50,000 offered the optimal balance between coverage and model efficiency. Smaller vocabularies (32K) increased unknown token frequency, while larger ones (64K+) showed diminishing returns.

    \item \textbf{Character Coverage}: Set to 0.9995 to ensure comprehensive coverage of Hindi's character set while excluding extremely rare Unicode points.

    \item \textbf{Special Tokens}: Added Hindi-specific markers beyond the standard set, including separation markers for compound verbs and noun phrases.

    \item \textbf{Normalization Rules}: Applied custom normalization rules for Hindi, including handling of nukta variations and homoglyph normalization.
\end{enumerate}

The training process included an 80-20 split between normalization rules and explicit character coverage, which I found produced more consistent tokenization across Hindi text variants.

\subsection{Tokenizer Evaluation}

To evaluate tokenizer quality, I developed several Hindi-specific metrics:

\begin{itemize}
    \item \textbf{Semantic Unit Preservation}: Percentage of semantic units preserved after tokenization and detokenization (92.7\%)
    \item \textbf{Morphological Segmentation Accuracy}: How well the tokenizer identifies meaningful morphological boundaries (87.3\%)
    \item \textbf{OOV Handling}: Processing of out-of-vocabulary words across test sets (4.2\% OOV rate)
    \item \textbf{Efficiency}: Average tokens per Hindi sentence compared to general tokenizers (22.4 vs 31.7 for multilingual tokenizers)
\end{itemize}

I was particularly satisfied with the tokenizer's handling of Hindi's rich morphology. For example, the compound word ``\hindiuniversity'' was properly segmented into ``\hindiworld'' and ``\hindischool'', preserving meaningful units, while multilingual tokenizers typically produced fragments that did not maintain semantic coherence.

\section{Model Architecture}
\label{sec:architecture}

With our custom tokenizer in place, I designed a transformer architecture specifically optimized for Hindi semantic representations.

\subsection{Core Architecture Design}

DeepRAG uses a modified transformer encoder with several Hindi-specific optimizations:

\begin{itemize}
    \item \textbf{Model Dimensionality}: 768-dimensional embeddings, providing sufficient capacity for Hindi semantic space while remaining computationally efficient
    
    \item \textbf{Depth and Width}: 12 transformer layers with 12 attention heads, determined through ablation studies as the optimal configuration for Hindi
    
    \item \textbf{Advanced Attention Mechanism}: Implemented rotary positional embeddings instead of standard positional encodings, which better captured Hindi's relatively free word order
    
    \item \textbf{Enhanced Feed-Forward}: Used SwiGLU activations instead of standard GELU for better gradient flow
    
    \item \textbf{Pre-Layer Normalization}: Applied layer normalization before each sub-layer rather than after, improving training stability
\end{itemize}

\subsection{Hindi-Specific Architectural Innovations}

Based on our corpus analysis, I incorporated several novel components:

\begin{enumerate}
    \item \textbf{Multi-Resolution Attention}: Added a mechanism to capture both character-level and word-level patterns simultaneously, which proved essential for Hindi's diverse orthographic conventions
    
    \item \textbf{Morphology-Aware Feed-Forward}: Modified feed-forward layers with additional projections targeting Hindi's morphological patterns
    
    \item \textbf{Script-Mix Processing}: Added specific handling for Hindi-English code-mixing, common in technical and modern texts
\end{enumerate}


\subsection{Pooling Strategy}

For sentence embedding models, the pooling strategy critically affects representation quality. I experimented with several approaches:

\begin{itemize}
    \item CLS token pooling (standard in BERT)
    \item Mean pooling (average of all token embeddings)
    \item Max pooling (element-wise maximum)
    \item Attention pooling (learned attention weights)
    \item Weighted pooling (our proposed approach)
\end{itemize}

Through extensive evaluation, I found that a novel weighted pooling strategy performed best for Hindi. This approach uses a learned weighting function that considers:

\begin{enumerate}
    \item Token position (with sensitivity to Hindi's SOV structure)
    \item Token importance (learned through attention mechanisms)
    \item Contextual significance (determined through layer-wise aggregation)
\end{enumerate}

This weighted pooling achieved a 9.3\% improvement in semantic similarity tasks compared to standard mean pooling.

The code for our weighted pooling implementation is shown below:

\begin{algorithm}
\caption{Weighted Pooling for Hindi}
\begin{algorithmic}[1]
\STATE \textbf{Input:} Token embeddings $E \in \mathbb{R}^{n \times d}$, attention mask $M \in \{0,1\}^n$
\STATE \textbf{Output:} Sentence embedding $S \in \mathbb{R}^d$
\STATE $W \gets \text{sigmoid}(\text{Linear}(E))$ \COMMENT{Token-wise weights}
\STATE $W \gets W \odot M$ \COMMENT{Apply mask to weights}
\STATE $W \gets \frac{W}{\sum W + \epsilon}$ \COMMENT{Normalize weights}
\STATE $S \gets \sum_i W_i \cdot E_i$ \COMMENT{Weighted sum of embeddings}
\STATE $S \gets \frac{S}{||S||_2}$ \COMMENT{L2 normalization}
\RETURN $S$
\end{algorithmic}
\end{algorithm}

\section{Training Methodology}
\label{sec:training}

\subsection{Dataset Creation}

Training high-quality embeddings requires carefully constructed datasets. I created a specialized dataset for Hindi semantic similarity through several mechanisms:

\begin{enumerate}
    \item \textbf{Parallel Sentences}: Extracted 500,000 pairs from parallel corpora with high semantic similarity
    
    \item \textbf{Hard Negatives}: Generated 300,000 challenging negative pairs with subtle semantic differences
    
    \item \textbf{Augmentation}: Created 250,000 pairs through controlled text augmentation to improve robustness
    
    \item \textbf{Synthetic Pairs}: Generated 450,000 pairs using larger Hindi LLMs to improve coverage
\end{enumerate}

Each pair was annotated with a similarity score in the [0,1] range. I used a combination of automated methods and manual verification to ensure quality.

\subsection{Loss Functions and Optimization}

I experimented with several loss functions for training:

\begin{itemize}
    \item Cosine similarity loss (baseline)
    \item Multiple negatives ranking loss
    \item Triplet loss
    \item InfoNCE contrastive loss
    \item Mixed similarity loss (our proposed approach)
\end{itemize}

The mixed similarity loss, combining aspects of MSE, contrastive, and triplet losses, proved most effective:

\begin{equation}
\mathcal{L} = \alpha \cdot \mathcal{L}_{\text{MSE}} + \beta \cdot \mathcal{L}_{\text{contrastive}} + \gamma \cdot \mathcal{L}_{\text{triplet}}
\end{equation}

With weights $\alpha=0.5$, $\beta=0.3$, and $\gamma=0.2$ determined through validation performance.

For optimization, I used AdamW with a learning rate of 2e-5, a cosine learning rate schedule with warmup, and gradient accumulation for stability.

\subsection{Training Infrastructure and Process}

The model was trained using distributed training across 4 NVIDIA A100 GPUs. Key training hyperparameters included:

\begin{itemize}
    \item Batch size: 128 per GPU (effective batch size 512)
    \item Training epochs: 10 (with early stopping)
    \item Gradient accumulation steps: 4
    \item Mixed precision: FP16
    \item Weight decay: 0.01
\end{itemize}

I implemented several training optimizations including:

\begin{enumerate}
    \item Gradient checkpointing to reduce memory usage
    \item Model parallelism for the largest model variants
    \item Dynamic batch sizing based on sequence length
    \item Curriculum learning, starting with easier examples
\end{enumerate}

During training, I monitored several key metrics including validation loss, semantic similarity correlation, and retrieval precision on a held-out test set. The model was selected based on retrieval performance rather than raw loss values, which I found better predicted real-world effectiveness.

\section{Evaluation and Results}
\label{sec:results}

\subsection{Intrinsic Evaluation}

I first evaluated DeepRAG on standard semantic similarity benchmarks:

\begin{table}[!t]
\caption{Semantic Similarity Performance (Spearman Correlation)}
\label{tab:semantic_sim}
\centering
\begin{tabular}{lcc}
\toprule
\textbf{Model} & \textbf{MHSS Dataset} & \textbf{InSTS Dataset} \\
\midrule
mBERT & 0.58 & 0.62 \\
XLM-R & 0.67 & 0.70 \\
LaBSE & 0.71 & 0.74 \\
mE5-base & 0.75 & 0.78 \\
mUSE & 0.70 & 0.72 \\
\midrule
DeepRAG-base & 0.81 & 0.83 \\
DeepRAG-large & \textbf{0.85} & \textbf{0.87} \\
\bottomrule
\end{tabular}
\end{table}

For these evaluations, I used:
\begin{itemize}
    \item MHSS (Multilingual Hindi Semantic Similarity) - a dataset of 2,500 Hindi sentence pairs with human similarity judgments
    \item InSTS (Indian Semantic Textual Similarity) - a collection of 1,800 Hindi sentence pairs with graded similarity scores
\end{itemize}

The results in Table \ref{tab:semantic_sim} demonstrate that DeepRAG significantly outperforms multilingual alternatives, with a 13.3\% relative improvement over the best competitor (mE5-base).

\subsection{Retrieval Evaluation}

More importantly for RAG applications, I evaluated retrieval performance using a custom Hindi document retrieval benchmark:

\begin{table}[!t]
\caption{Hindi Document Retrieval Performance}
\label{tab:retrieval}
\centering
\begin{tabular}{lccc}
\toprule
\textbf{Model} & \textbf{P@1} & \textbf{P@5} & \textbf{MRR} \\
\midrule
mBERT & 0.53 & 0.47 & 0.61 \\
XLM-R & 0.58 & 0.51 & 0.67 \\
LaBSE & 0.62 & 0.57 & 0.72 \\
mE5-base & 0.67 & 0.61 & 0.76 \\
mUSE & 0.60 & 0.53 & 0.70 \\
\midrule
DeepRAG-base & 0.79 & 0.72 & 0.86 \\
DeepRAG-large & \textbf{0.83} & \textbf{0.75} & \textbf{0.89} \\
\bottomrule
\end{tabular}
\end{table}

These experiments used a corpus of 100,000 Hindi documents with 1,000 queries requiring semantic understanding rather than simple keyword matching. The metrics evaluated were:

\begin{itemize}
    \item P@K: Precision at K (proportion of relevant documents in top K results)
    \item MRR: Mean Reciprocal Rank (average of reciprocal ranks of first relevant result)
\end{itemize}

DeepRAG demonstrated a 23.9\% improvement in P@1 over mE5-base, indicating substantially better retrieval precision for RAG applications.

\subsection{Ablation Studies}

To understand the contribution of each component, I conducted ablation studies removing key elements of DeepRAG:

\begin{table}[!t]
\caption{Ablation Study Results (MRR on Retrieval Task)}
\label{tab:ablation}
\centering
\begin{tabular}{lc}
\toprule
\textbf{Configuration} & \textbf{MRR} \\
\midrule
Full DeepRAG-base model & 0.86 \\
- Custom tokenizer (using multilingual) & 0.77 (-0.09) \\
- Weighted pooling (using mean) & 0.81 (-0.05) \\
- Mixed loss (using only MSE) & 0.80 (-0.06) \\
- Hindi-specific architecture & 0.79 (-0.07) \\
- All customizations (baseline transformer) & 0.68 (-0.18) \\
\bottomrule
\end{tabular}
\end{table}

These results highlight the significant contribution of our custom tokenizer, which alone accounts for approximately half of DeepRAG's performance advantage over baseline approaches.

\subsection{Qualitative Analysis}

Beyond quantitative metrics, I conducted qualitative analysis of retrieval examples. When examining cases where DeepRAG retrieved relevant documents that multilingual models missed, several patterns emerged:

\begin{enumerate}
    \item \textbf{Cultural Concepts}: DeepRAG better captured culture-specific concepts without direct English translations
    
    \item \textbf{Idiomatic Expressions}: Significantly better handling of Hindi idioms and figurative language
    
    \item \textbf{Syntactic Variations}: More robust to Hindi's flexible word order
    
    \item \textbf{Formal/Informal Distinctions}: Better matching across formal and colloquial variants of the same concept
\end{enumerate}

A particularly striking case involved retrieving documents about ``\hindiwaterconservation'', where DeepRAG correctly identified semantically relevant documents discussing ``\hindisavingwater'' and ``\hindiwatermanagement'', while multilingual models failed to establish these connections.

\section{RAG System Integration with LangChain}
\label{sec:rag_integration}

\subsection{LangChain Integration Architecture}

To demonstrate practical application, I integrated DeepRAG embeddings with LangChain to build complete Hindi RAG systems. Key integration components included:

\begin{enumerate}
    \item \textbf{Custom Embeddings Class}: Created a LangChain-compatible wrapper for DeepRAG embeddings
    
    \item \textbf{Text Chunking Strategies}: Developed Hindi-specific text chunking that respects sentence and paragraph boundaries
    
    \item \textbf{Vector Store Integration}: Connected DeepRAG with FAISS for efficient vector search
    
    \item \textbf{LLM Prompt Engineering}: Designed prompts for Hindi LLMs to effectively utilize retrieved context
\end{enumerate}

\subsection{Inference Optimization}

For production deployment, several optimizations were crucial:

\begin{itemize}
    \item \textbf{Model Quantization}: Applied 8-bit quantization, reducing model size by 75\% with minimal performance impact
    
    \item \textbf{Batched Inference}: Implemented efficient batching for document indexing
    
    \item \textbf{Caching Strategy}: Developed a two-level cache for frequently accessed embeddings
    
    \item \textbf{Result Re-ranking}: Added lightweight semantic re-ranking to improve precision
\end{itemize}

\subsection{End-to-End System Performance}

I evaluated complete RAG systems built with DeepRAG versus those using multilingual embeddings:

\begin{itemize}
    \item \textbf{Generation Quality}: 27\% improvement in human-judged answer quality
    
    \item \textbf{Factual Accuracy}: 18\% reduction in factual errors on a Hindi knowledge benchmark
    
    \item \textbf{Retrieval Efficiency}: 35\% improvement in processing speed due to more efficient tokenization and embedding
\end{itemize}

The complete LangChain integration allowed for rapid development of Hindi RAG applications, with significantly better performance than using off-the-shelf multilingual embedding models.

\section{Conclusion and Future Work}
\label{sec:conclusion}

In this paper, I've presented DeepRAG, a comprehensive approach to building custom Hindi embeddings for Retrieval Augmented Generation from scratch. By designing each component specifically for Hindi—from corpus collection and tokenization to model architecture and training methodology—DeepRAG achieves substantial improvements over multilingual alternatives. The model is available for public use at \url{https://huggingface.co/DeepMostInnovations/hindi-embedding-foundational-model}.

Key findings include:
\begin{itemize}
    \item The critical importance of language-specific tokenization for embedding quality
    \item The effectiveness of weighted pooling strategies for Hindi's linguistic features
    \item The significant performance gains from mixed similarity loss functions
    \item The practical benefits in RAG applications, with a 23\% retrieval precision improvement
\end{itemize}

These results clearly demonstrate that language-specific embedding models provide substantial benefits over multilingual approaches, particularly for languages like Hindi that are often under-represented in general-purpose models.

The field of Hindi NLP still has substantial room for growth. Future work could explore:

\begin{itemize}
    \item Expanding DeepRAG to other Indic languages while preserving language-specific optimizations
    \item Incorporating Hindi knowledge graphs to enhance semantic representations
    \item Developing specialized models for domains like legal, medical, or technical Hindi
    \item Creating instruction-tuned Hindi embeddings for more targeted retrieval
\end{itemize}

Personally, I believe that high-quality, language-specific embeddings are foundational infrastructure for bringing advanced NLP to non-English languages. DeepRAG contributes to this goal for Hindi, providing both practical tools and methodological insights applicable to other languages and domains.

My hope is that this work inspires similar efforts for other less-resourced languages, ultimately working toward more linguistically diverse and equitable NLP ecosystems.

\section*{Acknowledgment}
I would like to express my heartfelt gratitude to the open-source community and researchers who have created novel mathematical architectures and code bases for Large Language Models. Their tireless efforts, even through countless failed experiments and sleepless nights debugging mysterious model behaviors, have made this work possible. I'm particularly thankful to those who generously shared their knowledge on forums when I was stuck with perplexing tokenization errors at 3 AM, and to the anonymous contributors whose clever optimizations saved our training runs from crashing yet again. This work stands on the shoulders of brilliance, persistence, and countless cups of coffee. I must also acknowledge the patient colleagues who endured my excited ramblings about embedding spaces when they just wanted to eat lunch in peace. Despite my best efforts, any remaining errors or questionable design choices in this paper are entirely my own—though I secretly hope reviewers will be merciful in pointing them out.


\begin{thebibliography}{00}
\bibitem{gao2023hallucination}
S. Gao, T. Yao, and D. Chen, ``Hallucinations in large language models: A survey,'' \emph{arXiv preprint arXiv:2311.05232}, 2023.

\bibitem{wang2022text}
H. Wang, L. Xiong, and M. Rao, ``Text embeddings by weakly-supervised contrastive pre-training,'' \emph{arXiv preprint arXiv:2212.03533}, 2022.

\bibitem{feng2022language}
F. Feng, Y. Yang, D. Cer, N. Arivazhagan, and W. Wang, ``Language-agnostic BERT sentence embedding,'' \emph{Proceedings of the 60th Annual Meeting of the Association for Computational Linguistics}, pp. 878-891, 2022.

\bibitem{conneau2020unsupervised}
A. Conneau, K. Khandelwal, and N. Goyal, ``Unsupervised cross-lingual representation learning at scale,'' \emph{Proceedings of the 58th Annual Meeting of the Association for Computational Linguistics}, pp. 8440-8451, 2020.

\bibitem{song2020mpnet}
K. Song, X. Tan, T. Qin, J. Lu, and T.Y. Liu, ``MPNet: Masked and permuted pre-training for language understanding,'' \emph{Advances in Neural Information Processing Systems}, vol. 33, pp. 16857-16867, 2020.

\bibitem{xiao2023c}
L. Xiao, X. Zhao, and J. Chin, ``C-Pack: Packaged resources to advance general Chinese embedding,'' \emph{arXiv preprint arXiv:2309.07597}, 2023.

\bibitem{park2022klue}
S. Park, J. Moon, and S. Kim, ``KLUE: Korean language understanding evaluation,'' \emph{Transactions of the Association for Computational Linguistics}, vol. 10, pp. 652-670, 2022.

\bibitem{kakwani2020indicnlpsuite}
D. Kakwani, A. Kunchukuttan, and S. Golla, ``IndicNLPSuite: Monolingual corpora, evaluation benchmarks and pre-trained multilingual language models for Indian languages,'' \emph{Findings of the Association for Computational Linguistics: EMNLP 2020}, pp. 4948-4961, 2020.

\bibitem{kumar2022hindi}
R. Kumar, B. Lahiri, and A.K. Ojha, ``Developing resources and standardized evaluation for Hindi codemixing,'' \emph{Proceedings of the 13th Language Resources and Evaluation Conference}, pp. 3675-3685, 2022.

\bibitem{lewis2020retrieval}
P. Lewis, E. Perez, and A. Piktus, ``Retrieval-augmented generation for knowledge-intensive NLP tasks,'' \emph{Advances in Neural Information Processing Systems}, vol. 33, pp. 9459-9474, 2020.

\bibitem{gao2023retrieval}
L. Gao, X. Ma, J. Lin, and J. Callan, ``Precise zero-shot dense retrieval without relevance labels,'' \emph{Proceedings of the 46th International ACM SIGIR Conference on Research and Development in Information Retrieval}, pp. 2196-2206, 2023.

\bibitem{ruder2021unsupervised}
S. Ruder, M.E. Peters, S. Swayamdipta, and T. Wolf, ``Unsupervised cross-lingual representation learning,'' \emph{Journal of Artificial Intelligence Research}, vol. 71, pp. 363-392, 2021.

\bibitem{brants2006web}
T. Brants and A. Franz, ``Web 1T 5-gram Version 1,'' \emph{Linguistic Data Consortium}, 2006.

\bibitem{vaswani2017attention}
A. Vaswani, N. Shazeer, and N. Parmar, ``Attention is all you need,'' \emph{Advances in Neural Information Processing Systems}, vol. 30, pp. 5998-6008, 2017.

\bibitem{devlin2019bert}
J. Devlin, M.W. Chang, K. Lee, and K. Toutanova, ``BERT: Pre-training of deep bidirectional transformers for language understanding,'' \emph{Proceedings of the 2019 Conference of the North American Chapter of the Association for Computational Linguistics: Human Language Technologies}, pp. 4171-4186, 2019.

\bibitem{kudo2018sentencepiece}
T. Kudo and J. Richardson, ``SentencePiece: A simple and language independent subword tokenizer and detokenizer for neural text processing,'' \emph{Proceedings of the 2018 Conference on Empirical Methods in Natural Language Processing: System Demonstrations}, pp. 66-71, 2018.

\bibitem{langchain2023}
H. Chase, ``LangChain: Building applications with LLMs through composability,'' \url{https://github.com/hwchase17/langchain}, 2023.

\bibitem{roy2020machine}
D. Roy, D. Paul, M. Mitra, and U. Garain, ``Machine translation quality estimation for Indian languages: The IIT Bombay submission for WMT19 shared task,'' \emph{Proceedings of the Fifth Conference on Machine Translation}, pp. 873-879, 2020.

\bibitem{johnson2019mlqa}
J. Johnson, M. Douze, and H. Jégou, ``Billion-scale similarity search with GPUs,'' \emph{IEEE Transactions on Big Data}, vol. 7, no. 3, pp. 535-547, 2019.

\bibitem{joshi2017challenges}
P. Joshi, S. Santy, and A. Budhiraja, ``The state and fate of linguistic diversity and inclusion in the NLP world,'' \emph{Proceedings of the 58th Annual Meeting of the Association for Computational Linguistics}, pp. 6282-6293, 2020.

\bibitem{liu2018generating}
Y. Liu, M. Ott, and N. Goyal, ``RoBERTa: A robustly optimized BERT pretraining approach,'' \emph{arXiv preprint arXiv:1907.11692}, 2019.

\bibitem{le2020flaubert}
H. Le, L. Vial, and J. Frej, ``FlauBERT: Unsupervised language model pre-training for French,'' \emph{Proceedings of the 12th Language Resources and Evaluation Conference}, pp. 2479-2490, 2020.

\bibitem{martin2020camembert}
L. Martin, B. Muller, and P.J. Ortiz Suárez, ``CamemBERT: A tasty French language model,'' \emph{Proceedings of the 58th Annual Meeting of the Association for Computational Linguistics}, pp. 7203-7219, 2020.

\end{thebibliography}
\end{document}